\documentclass[runningheads]{llncs}

\usepackage{booktabs} 
\usepackage{amsmath,amssymb,amsfonts}
\usepackage{algorithm}
\usepackage{graphicx}
\usepackage{color}
\usepackage{array,ragged2e}
\usepackage[noend]{algpseudocode}
\usepackage{csquotes}
\usepackage{siunitx}
\usepackage[caption=false,font=normalsize,textfont=sf]{subfig}

\usepackage{titlesec}
\titlespacing*{\section}{0pt}{0.5\baselineskip}{0.5\baselineskip}
\titlespacing*{\subsection}{0pt}{0.5\baselineskip}{0.5\baselineskip}
\titlespacing*{\subsubsection}{0pt}{0.5\baselineskip}{0.5\baselineskip}

\DeclareMathOperator*{\argmax}{argmax}

\begin{document}
\title{Constructing Complexity-efficient Features in XCS with Tree-based Rule Conditions}
\titlerunning{Constructing Complexity-efficient Tree-based Features in XCS}

\author{Trung B. Nguyen\inst{1}\orcidID{0000-0002-1990-8647} \and
	Will N. Browne\inst{1}\orcidID{0000-0001-8979-2224} \and
	Mengjie Zhang\inst{1}\orcidID{0000-0003-4463-9538}}

\authorrunning{T. B. Nguyen et al.}

\institute{Victoria University of Wellington, Wellington 6140, NZ 
	\email{\{trung.nguyen,will.browne,mengjie.zhang\}@ecs.vuw.ac.nz}}

\maketitle

\begin{abstract}
A major goal of machine learning is to create techniques that abstract away irrelevant information. The generalisation property of standard Learning Classifier System (LCS) removes such information at the feature level but not at the feature interaction level. Code Fragments (CFs), a form of tree-based programs, introduced feature manipulation to discover important interactions, but they often contain irrelevant information, which causes structural inefficiency. XOF is a recently introduced LCS that uses CFs to encode building blocks of knowledge about feature interaction. This paper aims to optimise the structural efficiency of CFs in XOF. We propose two measures to improve constructing CFs to achieve this goal. Firstly, a new CF-fitness update estimates the applicability of CFs that also considers the structural complexity. The second measure we can use is a niche-based method of generating CFs. These approaches were tested on Even-parity and Hierarchical problems, which require highly complex combinations of input features to capture the data patterns. The results show that the proposed methods significantly increase the structural efficiency of CFs, which is estimated by the rule ``generality rate". This results in faster learning performance in the Hierarchical Majority-on problem. Furthermore, a user-set depth limit for CF generation is not needed as the learning agent will not adopt higher-level CFs once optimal CFs are constructed.

\end{abstract}

\section{Introduction}

A major goal of machine learning techniques is to abstract away irrelevant information. This improves the explainability of complex learned knowledge. A popular representation for encoding knowledge that encourages the explainability of learned knowledge is tree-based programs, such as Genetic Programming trees \cite{koza_genetic_1992}. However, the problem of bloat, i.e. meaningless or even harmful subtrees, in learned trees inhibits their explainability and hurts the performance of the system \cite{luke2006comparison}. It can be inefficient computationally and disrupts rule discovery by including poor building blocks of knowledge in recombination operators. In case of continual learning \cite{thrun2012learning} or layered learning \cite{stone2000layered}, the accumulated knowledge can suffer from exponentially increasing inefficiency in complex trees as the learning system continues to deal with more and more complex problems.

Learning Classifier Systems (LCSs) are a set of evolutionary techniques that enables layered learning \cite{alvarez2016human} and explainability due to their transparent and niche-based nature\footnote{Niche here means the local subsets of the data space} \cite{butz2006rule,urbanowicz_introduction_2017}. LCSs were originally a concept of cognitive systems that was adapted to become a rule-based system for machine learning and robotics, e.g. classification, regression, and multi-step navigation problems. XCS is a powerful Michigan-style LCS which complements the ``divide and conquer" ability inherent in LCSs with an accuracy-based fitness measure \cite{wilson_classifier_1995,butz_algorithmic_2000}. This enables XCS to divide a complex problem into subproblems with its niches to solve efficiently. XCS is also a framework where any complex representation can be integrated into its rules. Tree-based programs are the representation of interest as they enable higher-level feature construction that could encode ``abstract" data patterns. Trees also encourage the explainability of evolved rules as their structures can provide insights of the learned knowledge. A popularly used tree-based programs in XCS is Code Fragments (CFs), which have been introduced in XCS to improve its scalability \cite{iqbal2013reusing}.

\begin{figure}
	\vspace{-3mm}
	\centering
	\includegraphics[width=0.75\textwidth]{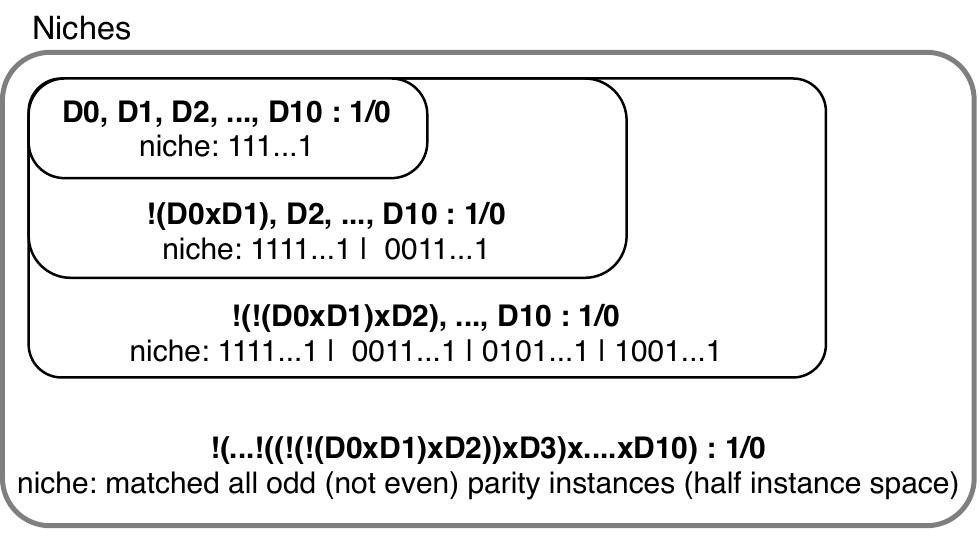}
	\caption{An example of generalising from a smallest niche to the largest one in a hierarchical problem (11-bit Even-parity problem) by XOF. $\times$ is the abbreviation for $XOR$. \enquote*{,}s separate CFs in a condition:action binary classifier effecting $1$ or $0$. $!$ is $NOT$. The generalising process benefits from growing from lowest-level CFs ($D_0,D_1,...,D_{10}$) to more complex ones gradually, i.e. $(D_0\times D_1)$, $(!(D_0\times D_1)\times D_2)$, ..., $(!(...!(!(!(D_0\times D_1)\times D_2)\times D_3)\times...\times D_{10}))$.}
	\label{fig:generalise}
	\vspace{-3mm}
\end{figure}

XOF is one such system that grows high-level CFs based on a set of the most applicable CFs that are included in an Observed List (OL) \cite{nguyen2019online}. XOF can learn hierarchical and large-scale problems by capturing the data patterns in CFs. However, its constructed trees also contain bloat due to the panmictic crossover of CFs in the OL. The general learning process of an XOF, when addressing a hierarchical problem, is to generalise from small niches, i.e. some specific cases, to larger niches by combining the building blocks from the small niches. Figure \ref{fig:generalise} illustrates the general relationship of the lower-level CFs in less generalised rules and the higher-level CFs in more generalised rules that can replace all the more specific rules. The higher-level CFs here is shown to combine the lower-level CFs to create CFs that can describe a superset niche. This heuristic suggests that a niching method for CFs can be beneficial for the generalising process of XOF.


Niching is a unique advantage of XCS. Previous implementations of XOF have not included any niching property for constructing CFs. This means that all CFs in the OL and in the CF population are grouped together without any discrimination among niches. If the information that a CF in the OL performs the best in the current niche or another niche is available, the learning system can avoid combing CFs from unrelated niches, which was the likely cause of the in non-optimal trees with bloat.


This paper proposes two novel methods to combat bloat in CFs. Accordingly, the objectives of this paper are as follows:

\begin{enumerate}
	\item To develop a new CF-fitness measure, which is to estimate the applicability of CFs in creating high-fitness classifiers, to emphasise the efficiency of CF structures. We also apply criteria based on the structural efficiency of CFs to select the most applicable CFs accordingly.
	\item To introduce a niching method for the CFs in the OL.
	\item To investigate the influence of the two implemented approaches on the structural efficiency of constructed CFs and the learning performances of XOF.
\end{enumerate}

The system will be tested on complex problems that require building hierarchical features to capture the patterns of data. Benchmark problems include Even-parity, Hierarchical Multiplexer, and Hierarchical Majority-on problems \cite{butz2006rule} because these Boolean problems require accurate hierarchical combinations of input attributes. These combinations must match with the data patterns of these problems to carry the maximal discriminative information of the problems. As a result, constructing such combinations can reduce the search space of rules in XOF. However, finding accurate complex combinations in CFs is challenging due to the large search space. Even-parity problems stress generalisation ability, which is not applicable to standard XCS using the ternary alphabet in rule conditions. In addition to posing the generalisation challenge, the Hierarchical Multiplexer domain is also epistatic and heterogeneous, while Hierarchical Majority-on domain has overlapping niches.



\section{Background}

\subsection{Learning Classifier Systems}
LCSs are a family of rule-based algorithms based on a concept of cognitive system \cite{holland_adaptation_1975}. An LCS generally interacts with an environment representing a target problem to evolve a population of rules using evolutionary techniques \cite{urbanowicz_introduction_2017}. XCS is a simplified reinforcement learning implementation of LCS that can be easily adapted to machine learning problems, robotic tasks, etc. \cite{butz_algorithmic_2000,wilson_classifier_1995}. The rule conditions of XCS enable it to ``divide" a hard problem into subproblems, i.e. niches, and ``conquer" each niche more easily. 



\subsection{CF-based XCSs}\label{ssect:CFXCSs}

Code Fragments (CFs) was originally introduced as binary tree-based programs used in Boolean problems with a depth limit of $2$ \cite{iqbal2013reusing}. A CF tree is connected graph with internal nodes corresponding to functions from a function set and terminal/leaf nodes representing input attributes from the environment state or reused learned CFs. The function set for Boolean domains usually has general binary operators, such as $\{AND, OR, NOT, NAND, XOR\}$.

CFs can describe complex patterns of data which enable XCS to generalise its rules and solve hierarchical problems using more compacted rule-sets. CFs can be used to represent rule conditions or actions in XCS \cite{iqbal_evolving_2013,iqbal2013reusing}. However, the power of tree-based programs also comes with challenges, especially the large search space in search of trees with high discriminative information for a problem. Existing methods used transfer learning and layered learning to resolve this issue. Iqbal et al. introduced XCSCFC that transfers learned CFs to the leaf nodes of new CFs in rule conditions to scale up to 135-bit Multiplexer problem \cite{iqbal2013reusing}. XCSCF\textsuperscript{2} extended the transferring capability of CFs to function nodes. This method used the rule populations of solved problems as rule-set functions for function nodes \cite{alvarez2014reusing}. Based on reusing rule-set functions, XCSCF* solved the general Multiplexer problem by decomposing the problem domain into subproblems and combining the solutions of the subproblems \cite{alvarez2016human}. These approaches require human guidance for either the transferring process or the learning order.

\subsection{XOF}

\begin{figure}
	\vspace{-3mm}
	\centering
	\includegraphics[width=0.7\textwidth]{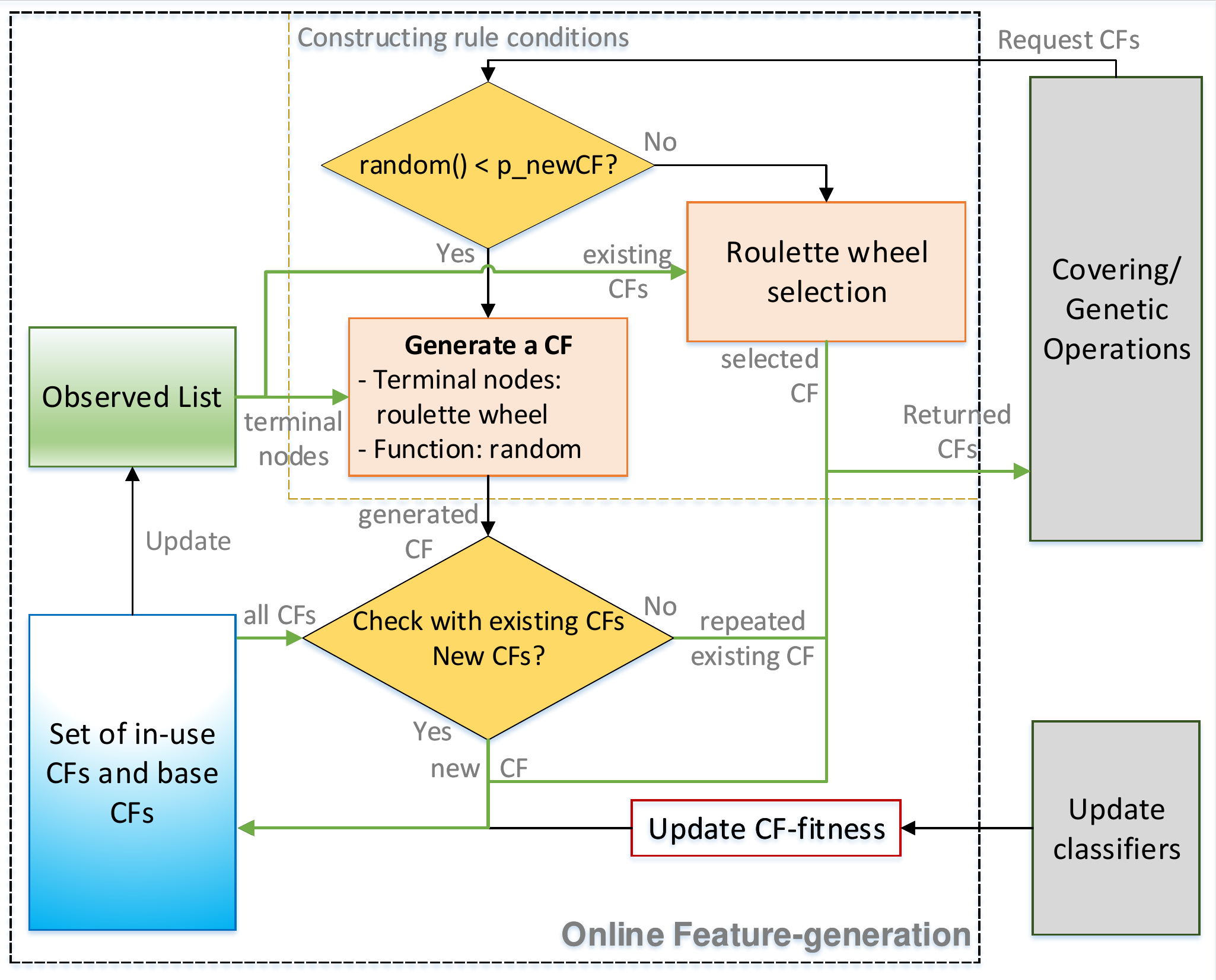}
	\caption{The OF module in XOF. It returns a CF when being requested by general processes of XCS. The returned CF can be an existing one selected by Roulette Wheel selection or a newly generated CF. In both cases, the OF relies on the OL.}
	\label{fig:OF}
	\vspace{-3mm}
\end{figure}

Similar to XCSCFC, an XCS with CFs in rule conditions, XOF grows from the initial data features to learn deeper tree-based features (CFs) containing discriminative patterns for classification problems \cite{nguyen2019improvement,nguyen2019online}. XOF extended XCS with the Online Feature-generation (OF) module (see Figure \ref{fig:OF}) and introduced CF-fitness as a parameter to estimate the applicability of CFs. The applicability of a CF is defined as its capability of producing high-fitness classifiers using the CF in the condition parts.

The OF constructs tree-based features by combining the most useful CFs in the Observed List (OL) to grow higher-level useful CFs for rule conditions. The ability of CFs to create more accurate and generalised classifiers defines CF-fitness (the applicability of CFs):

\vspace{-2mm}
\begin{equation}
	cf.f=\frac{cl.f}{\text{number of CFs in $cl$}},
\end{equation}
where $cf.f$ is the CF-fitness of CF $cf$, and $cl.f$ is the fitness of classifier $cl$. This applicability measurement evaluates a CF according to the accuracy and generality per CF of the highest-fitness classifier containing the CF. Thus, this CF-fitness creates pressure in combining CFs in the conditions of high-fitness classifiers without caring about the rule complexity. The combination pushed by this CF-fitness may result in new classifiers without increased generality. 

The evolutions of rules and CFs can mutually support each other. The rule fitness guides the evolution of CFs, while the evolution of CFs provides building blocks for rule conditions that contain more discriminative information of the patterns of data. In the first versions of XOFs, the learning system periodically updates the OL every $500$ iterations using the tournament selection based on CF-fitness. XOF-BF, the baseline algorithm in this paper, updates the CF-fitness of a CF according to its best classifier. The best classifier here refers to the highest-fitness classifier containing the CF. Because XOF enables learning of high-level trees, it also contains irrelevant information in the form of bloat.

\section{Method}

\subsection{Generalising CF-fitness using Rule-Fitness Rate}
\label{ss:cff_fitness_rate}

To improve the structural efficiency of CFs, we first introduce the term ``complexity" of CFs, which estimates the structural complexity of CFs. Because CFs here are binary trees (without considering negation, i.e. function $NOT$, as a separated node and adding complexity \cite{nguyen2019online}), the number of function nodes (internal nodes) is always $1$ less than the number of leaf nodes. Thus, we define the complexity of a CF as the number of leaf nodes, which are the amount of input information involved in evaluating the CF. Accordingly, the complexity of a rule is the accumulated complexity of all CFs in its condition. 

According to \cite{nguyen2019improvement}, the CF-based XCS targets to generate highly applicable tree-based features. Therefore, the structural efficiency of a CF is its capability to construct high-fitness rules with the least rule complexity. We use ``fitness rate'' of a classifier, which is equivalent to the fitness per unit of classifier complexity:

\begin{equation}
cl.f\_rate=\frac{cl.f}{cl.complexity},
\end{equation}
to estimate the CF-fitness of all CFs within the classifier. The CF-fitness of a CF based on the rule-fitness rate of the classifier (containing the CF) rewards higher CF-fitness on the CF that can construct accurate and more generalised rules using the least input information. Thus, we call this CF-fitness as Generalising CF-fitness (GCFF). At this point, this CF-fitness has similar goals with the rule fitness of traditional XCS. The additional benefit is that CF-based conditions enable more complex patterns with the same input attributes compared with XCS, which can result in larger niches.

Accordingly, the OL gathers CFs from the conditions of the classifiers with the highest fitness per complexity unit in the action set of XCS (see Section \ref{ss:simplified}). The updates of CF-fitness also follow the Widrow-Hoff learning rule \cite{sutton1988learning} based on the classifier with the highest fitness per complexity unit containing the CF:

\vspace{-2mm}
\begin{equation}
cf.f\mathrel{{+}{=}}\beta_{cf}*(\max_{cl \mid cf \in cl.condition} \frac{cl.f}{cl.complexity} - cf.f),
\end{equation}
where $\beta_{cf}$ is the learning rate of CF-fitness \cite{nguyen2019online}. This CF-fitness represents the highest fitness per complexity unit of a rule among rules having the CF, therefore called rule-fitness rate. In short, the OF module evaluates generated CFs based on their efficiency of CFs in using binary functions to combine the input attributes to produce accurate and generalised classifiers. 

\subsection{Niching for CFs}
\label{ss:niching_cfs}

The part of the OF module that associates with the learning processes of XCS, i.e. covering and genetic operations, is the CF generation by either selecting an existing CF or constructing a new one. As this process relies on CF-fitness, a niching method for CFs needs to be implemented for at least the CF-fitness. We develop a niching method that calibrates the CF-fitness of a CF based on the performance of the CF on the current niche. This method is designed to create boundaries between niches to prevent continual undesirable sharing of CFs. While sharing knowledge among niches is generally beneficial in many problems, undesirable transfers of CFs between niches can hold back the discovery of optimal building blocks for each niche.

The niching method calibrates the CF-fitness in three cases to estimate a local CF-fitness for the CF. First, if a CF has its best classifier matched in the current action set, this CF is known to perform the best in this niche. In this case, the OF uses its CF-fitness directly. The second case is when a CF never appears in any classifier in the current action set. The system obviously has no data on its actual performance in this niche. This niching method estimates the local CF-fitness of this CF naively with a constant rate of $0.1$ of its global CF-fitness. This value should be further investigated. The third case is in the middle of the first two cases when a CF does appear in at least one classifier in this local niche, but its best classifier is not the best overall. The estimated local CF-fitness of this CF is as follows:
\begin{equation}
	cf.f_{local} = cf.f * \frac{cf.local\_best\_classifier.f}{cf.best\_classifier.f},
	\label{eq:local_fitness}
\end{equation}
where the $cf.local\_best\_classifier$ is the ``best" classifier containing the CF in the current action set $[A]$, and the $cf.best\_classifier$ is its global ``best" classifier in the whole rule population $[P]$. It is noted that the definition of classifier being the ``best" for a CF varies according to the CF-fitness. In XOF-BF, it is the highest-fitness classifier containing the CF. Because we will test this niching method with the implementation that stacks this method with generalising CF-fitness, the quality of classifier is based on this new CF-fitness. Specifically, classifiers selected for Equation \ref{eq:local_fitness} are the ones with the highest rule-fitness rate:
\begin{align}
&cf.local\_best\_classifier = \argmax_{cl \in [A] \mid cf \in cl.condition} cl.f/cl.complexity,\\
&cf.best\_classifier = \argmax_{cl \in [P] \mid cf \in cl.condition} cl.f/cl.complexity,
\end{align}

\subsection{The Simplified OL Update}
\label{ss:simplified}

In this work, we simplify XOF's processes and thereby eliminate a number of hyper-parameters. Also, the remaining hyper-parameters can still control the pace of the evolution of CFs, such as the learning rate for CF-fitness $\beta_{cf}$. Instead of periodical updates, the system updates the OL in every exploiting iteration with two processes. The first process is to collect the CFs in the conditions of the classifiers that best represent the action sets. For example, when using the CF-fitness in Section \ref{ss:cff_fitness_rate}, the classifiers satisfying the following criteria will be used to collect the CFs for the OL:

\begin{equation}
cl.f/cl.complexity\geqslant 0.9*\max_{cl \in [A]}cl.f/cl.complexity,
\label{eq:OL_selection}
\end{equation}
where $0.9$ represents the selectivity of the OL. This value is empirically chosen among high values to compress the OL size. The second process is to remove CFs in outdated classifiers in the current niche that do not satisfy Eq. \ref{eq:OL_selection}. This step could remove necessary building blocks of other niches in case of the problems with overlapping niches. However, as we place the OL update before genetic operations, the removed necessary CFs in other niches are always added back. This method is to collect all necessary building blocks in all niches.


\section{Experiments}

\subsection{Generality Rate to Estimate the Structural Efficiency of CFs}

To compare the ability to generate complexity-efficient CFs, we tracked and evaluated the structural efficiency of the CFs in the highest-fitness classifiers. Also, the evaluation should be the least niche-biased, which focuses more on a subset of niches. Thus, the classifiers for collecting CFs for tracking the structural efficiency are gathered from at most one classifier per action set. These classifiers also need to be accurate and experienced to avoid irrelevant estimation of performance, such as the high structural efficiency of an inaccurate general rule $cl.experience\geq \theta_{GA}$ and $cl.error\leq \epsilon_0$ \cite{butz_algorithmic_2000}. 

This estimation of structural efficiency is still somewhat niche-biased because any niche with no experienced and accurate classifiers has no contribution to the estimated structural efficiency. This case is common when the accuracy is not $100\%$, but does not occur otherwise. Even after achieving $100\%$ accuracy, the estimation of the CF-structural efficiency can still be niche-biased if the estimation is not weighted by niche size. However, precise measurement requires that niche sizes that are prerequisite knowledge for a given problem. As we try to be naive about the tested problems, the evaluation will approximate the evolution of structural efficiency of the highest-fitness classifiers by averaging them among niches where experienced and accurate classifiers are available.

Having the representative classifiers to collect the most applicable CFs of the tested problem, we need a method to estimate the structural efficiency of these CFs. Since these CFs are from experienced and accurate classifiers, the other aspect of efficiency is only the generality \cite{nguyen2019online}. Therefore, the structural efficiency should involve the generality and complexity. We track the ``generality rate" of these classifiers to evaluate the structural efficiency of a classifier:

\vspace{-3mm}
\begin{align*}
&cl.generality=\frac{cl.matches}{cl.matches+cl.no\_matches}, \text{ thus}\\
&cl.generality\_rate=\frac{cl.generality}{cl.complexity},
\end{align*}
where $cl.matches$ and $cl.no\_matches$ respectively track the numbers of times classifier $cl$ matches and does not match all instances since it was created, and therefore the part $cl.matches/(cl.matches+cl.no\_matches)$ provides the generality of classifier $cl$ as it tracks the probability that classifier $cl$ matches any instance. The generality rate of classifiers estimates their efficiency in using $cl.complexity$ complexity units (CF structures) to produce accurate classifiers with the highest generality.

\subsection{Experimental Design}

In this section, we compare XOF with the two newly implemented features with itself in the lack of one or both new features. There are two criteria for comparisons: the learning performance as well as the discovery of complexity-efficient CFs. All experiments were evaluated based on the average of $30$ runs using $30$ independent random seeds. In this paper, we abbreviate the existing CF-fitness focusing on shortening rule conditions as Shortening CF-fitness (SCFF) and the niching method for CFs as Niching CFs (NCF). Thus, besides the existing version named XOF-BF \cite{nguyen2019improvement}, we will experiment with three new other approaches abbreviated as XOF-SCFF, XOF-GCFF, and XOF-GCFF-NCF within this paper. All of these new systems use the simplified OL update in Section \ref{ss:simplified}.

We configured all parameters of all tested versions of XOF equally except for the rule population size and stopping iteration. A general configuration of XOF \cite{nguyen2019improvement} is used in these experiments: the learning rate for rule parameters $\beta=0.2$ and the learning rate for CFs $\beta_{cf}=0.001$; the crossover rate is $\chi=0.2$; the mutation rate is $\mu=0.9$; the experience thresholds for deletion is $\theta_{del}=20$; the initial fitness of covered classifiers are $F_{init}=0.01$; the probability of specificness $p_{spec}=0.25$ with maximum rule-condition length set at twice the number of original input attributes; and the experience thresholds for subsumption $\theta_{sub}=50$. All three newly implemented versions of XOF have no limit on the OL size and a redundantly high limit on the CF depth, i.e. the maximum depth is $20$.

\subsection{Results on 11-bit Even-parity Problem}

\begin{figure}
	\vspace{-3mm}
	\centering
	\subfloat[Accuracy]{
		\label{fig:res_epar:acc}		
		\includegraphics[width=0.5\textwidth]{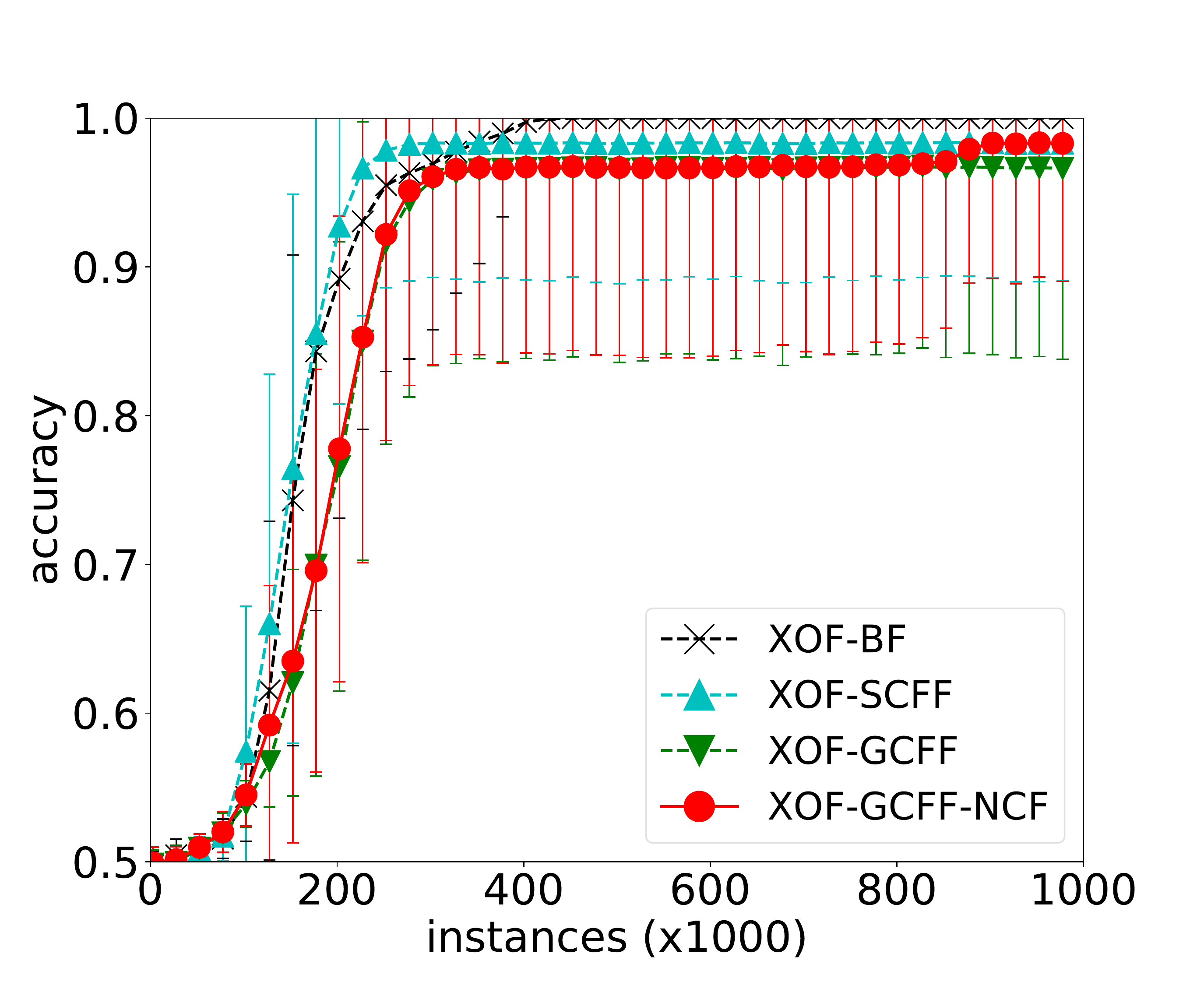}}
	\subfloat[Generality Rate]{
		\label{fig:res_epar:genex}		
		\includegraphics[width=0.5\textwidth]{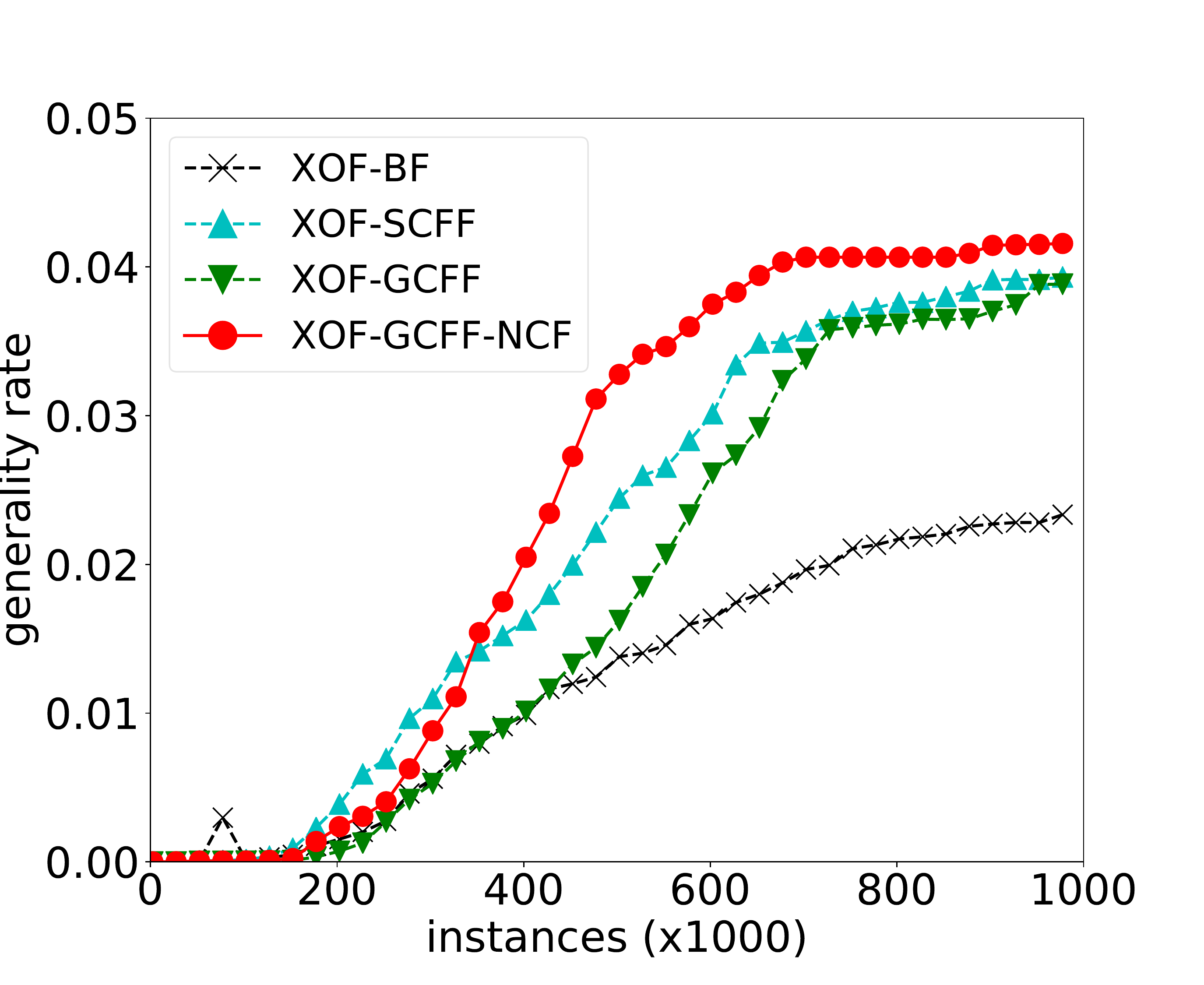}}
	\caption{Results on 11-bit Even-parity problem.}
	\label{fig:res_epar}
	\vspace{-3mm}
\end{figure}

The population sizes for all systems on this problem were equal at $8000$ classifiers. The learning graphs of tested approaches in this experiment were not substantially different except for the convergence phase, see Figure \ref{fig:res_epar:acc}. XOF-SCFF and XOF-BF had a slight advantage in the early phase. In $30$ runs of the three new systems in this paper, there was always one or two runs that were stuck at $50\%$ accuracy. The reason for being stuck was that the evolution of CFs creates an extra force to push the evolution of rules further to the local optima. Also, this Even-parity problem already poses a high probability of local optima for XCS as the probability of finding correct rules in XCS is very low. All inaccurate rules have the same accuracy of $50\%$, including the simplest rules and the rules with genotypes near the accurate ones. These stuck runs always ended up with the domination of a few very short rules (with only one CF in its conditions). The high rule numerosity and simple rule conditions caused the CFs in these rules to achieve higher CF-fitness and thereafter pushed these rules to earn more numerosity through genetic operations.

All three systems discovered CFs with significantly more efficient structures than ones by XOF-BF. The system with the niching method for CFs evolved the most optimal CFs (Figure \ref{fig:res_epar:genex}), although there were two stuck runs in most of the tested experience (\num[group-separator={,}]{1000000} instances), which had very low generality rate. The evolved CFs of this system reached near the optimal generality rate for this problem. The optimal generality rate is equivalent to the generality rate of the most generalised classifier in Figure \ref{fig:generalise}. This classifier has the generality of $0.5$ as it matches half of all instances. Also, its complexity is $11$ for $11$ leaf nodes. Hence, its generality rate is $0.5/11=0.04545$, which is the optimal generality rate here.

\subsection{Results on Hierarchical Problems}

We use 18-bit Hierarchical Multiplexer and 18-bit Hierarchical Majority-on problems to evaluate the generality rate of the tested approaches. These two problems pose relatively large search spaces as their hierarchy adds complexity to the data patterns. To capture these complex patterns, constructed CFs need to cover all $6$ non-overlapped three-successive-bit chunks with 3-bit Even-parity problems. An optimal CF that can cover a chunk, say $(D_3,D_4,D_5)$, has to use $XOR$ in all function nodes except for any arbitrary negation, such as $(!((!D_3)\times D_4))\times D_5)$. Such CFs can match half of all possibilities for the three bits and the other half by its negated version, which is not counted as a distinct construction in XOF because each CF has one corresponding negated version.

\begin{figure}
	\vspace{-3mm}
	\centering
	\subfloat[Accuracy]{
		\label{fig:res_hmux18:acc}		
		\includegraphics[width=0.49\textwidth]{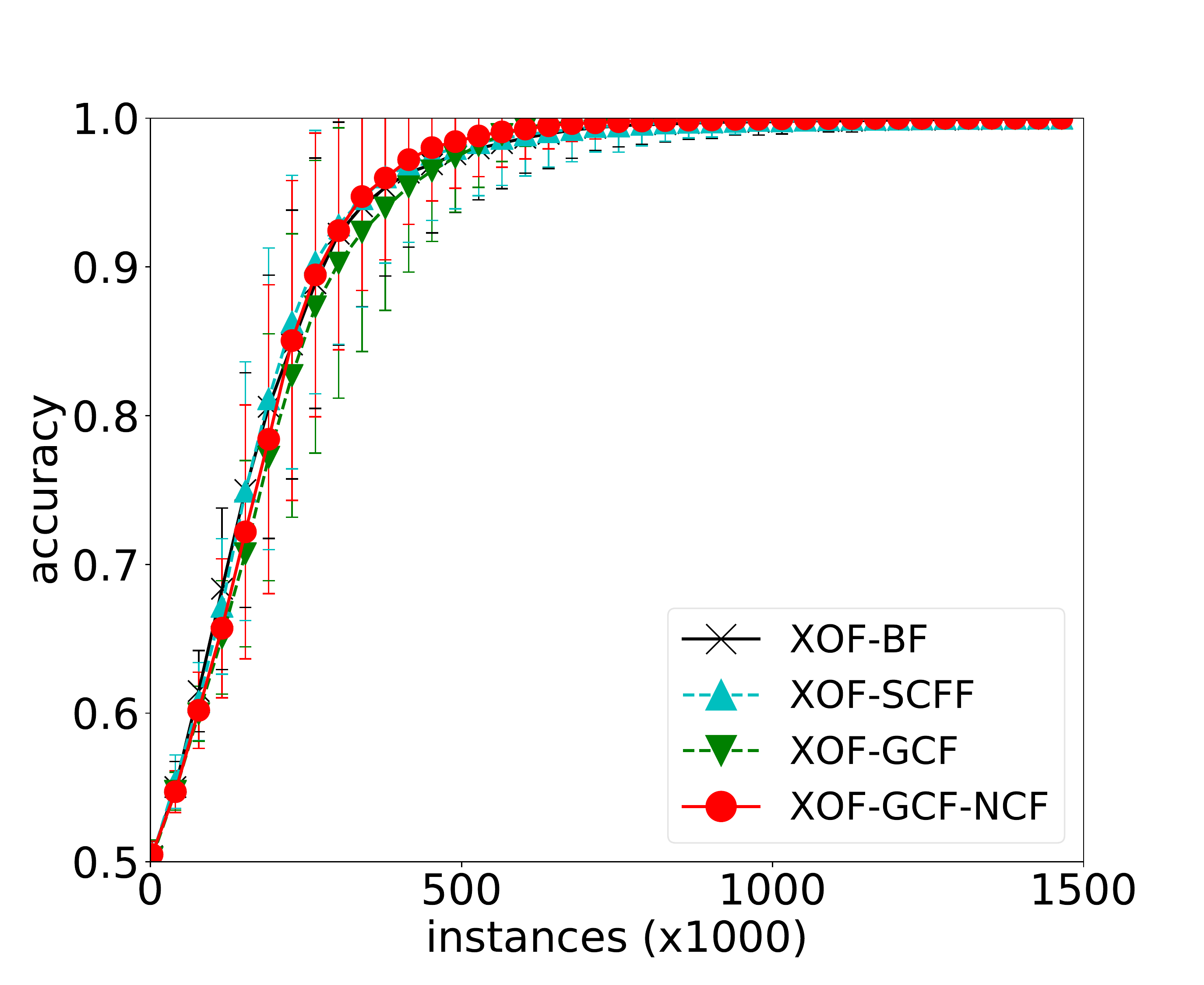}}
	\hfill
	\subfloat[Generality Rate]{
		\label{fig:res_hmux18:genex}		
		\includegraphics[width=0.49\textwidth]{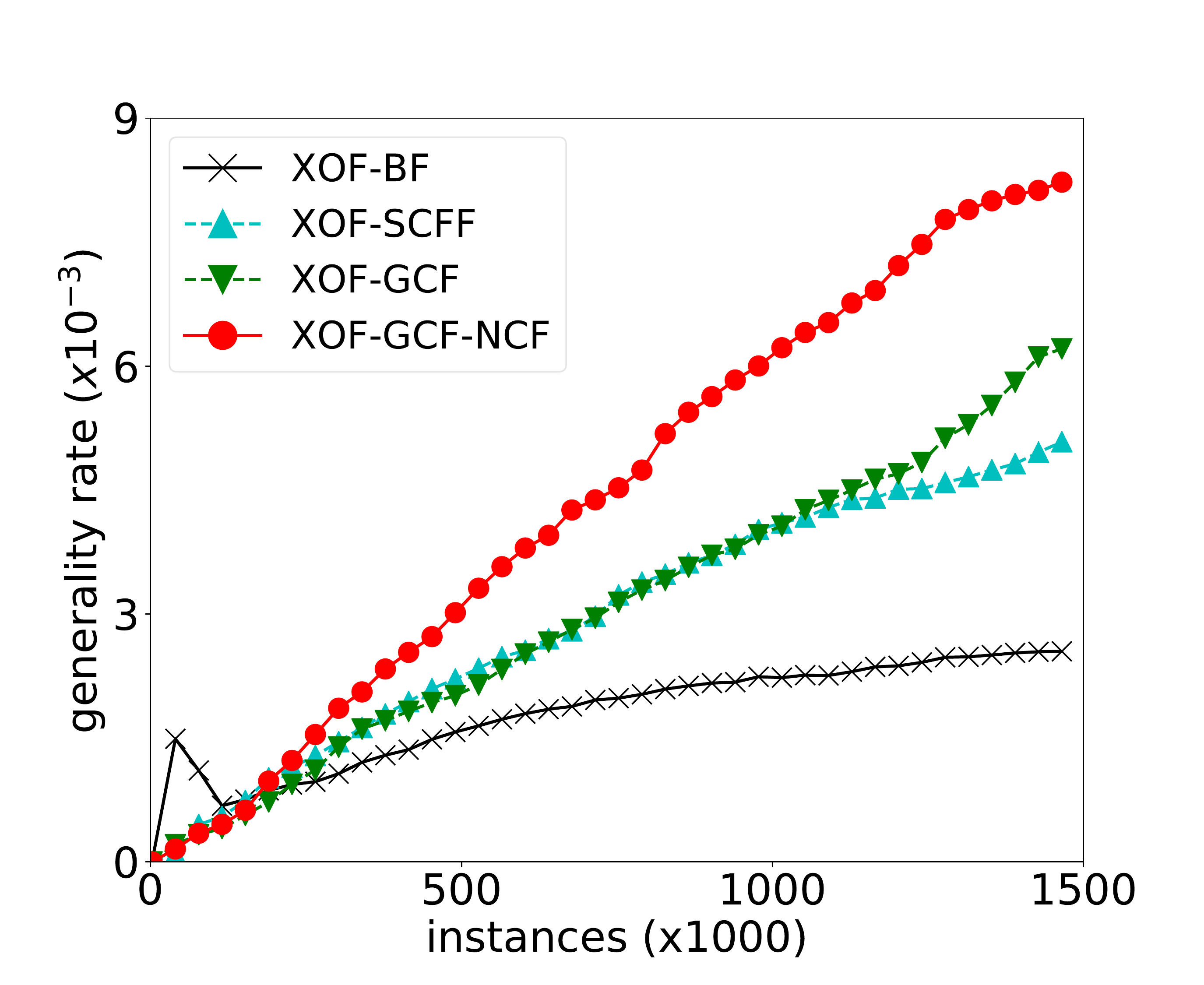}}
	\caption{Results on 18-bit Hierarchical Multiplexer problem.}
	\label{fig:res_hmux18}
	\vspace{-3mm}
\end{figure}

\begin{figure}
	\vspace{-3mm}
	\centering
	\subfloat[Accuracy]{
		\label{fig:res_hmaj18:acc}		
		\includegraphics[width=0.49\textwidth]{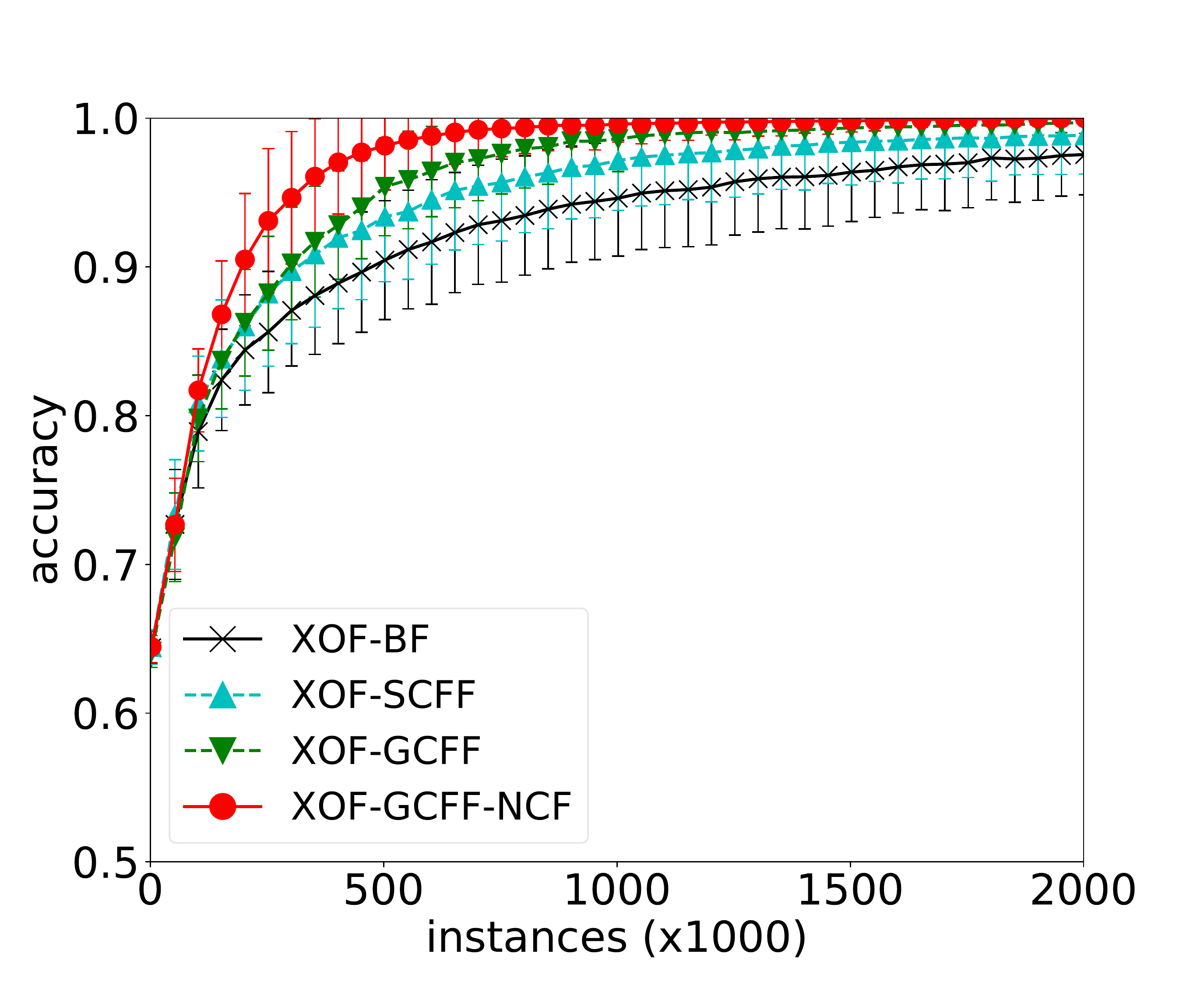}}
	\hfill
	\subfloat[Generality Rate]{
		\label{fig:res_hmaj18:genex}		
		\includegraphics[width=0.49\textwidth]{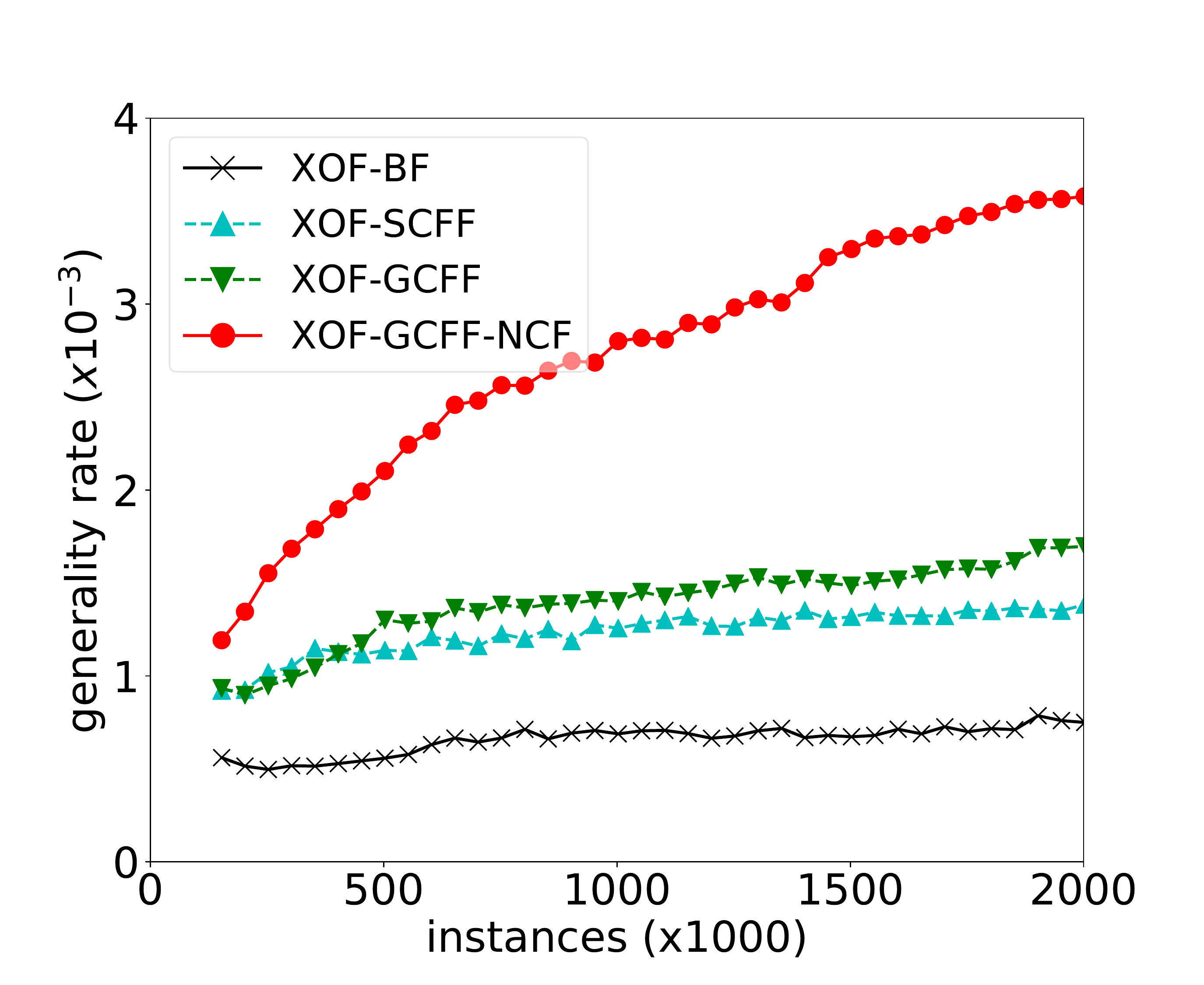}}
	\caption{Results on 18-bit Hierarchical Majority-on problem. The beginning parts of generality rates are omitted because over-general (inaccurate) rules that are temporarily accurate and experienced can create unreliable estimation.}
	\label{fig:res_hmaj18}
	\vspace{-3mm}
\end{figure}

All systems in these two experiments had the same population size of \num[group-separator={,}]{20000}. The learning performances of all approaches are no substantially different from one another in the 18-bit Hierarchical Multiplexer problem, see Figure \ref{fig:res_hmux18:acc}. In the 18-bit Hierarchical Majority-on problem, two XOFs with the SCFF, XOF-SCFF and XOF-BF, did not converge to $100\%$ accuracy, while the two XOFs using the GCFF, XOF-GCFF and XOF-GCFF-NCF, did (see Figure \ref{fig:res_hmaj18:acc}). The niching CFs also improves the learning performances of XOF in both experiments.

In these two experiments, XOF with niching CFs (XOF-GCFF-NCF) yielded the most optimal CFs among the tested versions of XOF (Figures \ref{fig:res_hmux18:genex} and \ref{fig:res_hmaj18:genex}). The evolution of CFs in XOF-BF again was stuck, plus it contained the most bloated CFs. Table \ref{tab:OL_samples} illustrates a few samples of CFs in the OLs of three tested systems. We show CFs related to chunks without optimal CFs discovered to show the difference of the influence by the two CF-fitness. SCFF, the CF-fitness focusing on shortening rule conditions, rated the two non-optimal CFs $(!((!D_{14})\times D_{13}))\land D_{12}$ and $(!((!D_{14})\times D_{13}))\lor D_{12}$ much higher CF-fitness than the lower-level CFs $!((!D_{14})\times D_{13})$ and $D_{12}$. The two latter CFs were the ones that can be combined to construct optimal CFs for the 3-bit Even-parity problem (on three bits $(D_{12},D_{13},D_{14})$). Although the patterns of the two former CFs did not generalise more than the lower-level ones on the 3-bit Even-parity problem, these higher-level CFs still had higher CF-fitness because they can produce rules with shorter conditions and the same patterns. This process even hindered discovering the optimal CFs for this 3-bit Even-parity chunk because the higher-level CFs have more probabilities to be selected in constructing CFs. Meanwhile, GCFF, the generalising CF-fitness, did not face this problem because such non-optimal combinations do not achieve higher CF-fitness than the lower-level CFs.

\begin{table}
	\vspace{-3mm}
	\centering
	\caption{A few samples of CFs with their CF-fitness in the OLs of XOF-SCFF, XOF-GCFF, and XOF-GCFF-NCF after \num[group-separator={,}]{150000} instances of learning the 18-bit Hierarchical Multiplexer problem. These samples were CFs related to a chunk that optimal CFs have not been constructed. To be as fair as possible, these samples were chosen from runs with generality rates in the small range from $0.0075$ to $0.0080$.}
	\begin{tabular}{ |m{45mm}|m{40mm}|m{32mm}| }
		\hline
		\Centering{\textbf{XOF-SCFF}} & \Centering{\textbf{XOF-GCFF}} & \Centering{\textbf{XOF-GCFF-NCF}} \\ \hline
		$(!((!D_{14})\times D_{13}))\land D_{12}$ $(0.157)$ & $(!D_9)\land(!D_{11})$ $(0.058)$ & $(!D_8)\times (!D_7)$ $(0.071)$ \\
		$(!((!D_{14})\times D_{13}))\lor D_{12}$ $(0.157)$ & $(D_9\land D_{11})\times (!D_{10})$ $(0.056)$ & $(!D_8)\times D_7$ $(0.061)$ \\ 
		$!((!D_{14})\times D_{13})$ $(0.121)$ & $D_{10}$ $(0.057)$ & $D_6$ $(0.071)$ \\ 
		$D_{12}$ $(0.121)$ & $D_{11}$ $(0.054)$ & n/a  \\
		n/a & $D_9$ $(0.054)$ & n/a \\ \hline
	\end{tabular}
	\label{tab:OL_samples}
	\vspace{-3mm}
\end{table}


\section{Further Discussions}

SCFF rewards higher CF-fitness on complex patterns without adding any more generalisation. This mechanism is likely to push the evolution of tree features to early local optima. The new CF-fitness, GCFF, only rewards higher CF-fitness on more complex patterns when these patterns can construct more generalised rules. Thus, the depth growth in the system using generalising CF-fitness will be slower and more reliable. As a result, the system can avoid being trapped in local optima with non-optimal CFs.

On the 11-bit Even-parity problem, the pure pressure on combining CFs and shortening rule conditions of SCFF has slightly better learning performance than the generalising pressure of GCFF. This can be explained as this CF-fitness estimation awards higher CF-fitness on combined CFs, which pushes the generalisation process faster. However, GCFF has a better performance on the Hierarchical Majority-on problem because it does not push the system towards rules with shorter conditions, which can easily become over-general rules in problems with overlapping niches.

The niching method for CFs improves the structural efficiency of CFs in XOF significantly, as shown by its superior average generality rate. Niching CFs guides combining optimal CFs to generalise existing patterns. Therefore, the evolution of CFs with niching CFs is accelerated without adding the likelihood of being trapped in local optima. This also results in slightly faster learning performances on the tested Hierarchical problems.

\section{Conclusions}

We have developed a new CF-fitness, called generalising CF-fitness, that focuses on more generalised patterns and avoid naively combining existing CFs. Accordingly, other processes of XOF have been adjusted to select CFs following the new criteria. The new CF-fitness slows down the growth of CF depth but adds more reliability to the CF construction. Although the structural efficiency of generated CFs has not been improved, it enables integrating a newly developed niching method for CFs, which results in accelerating the evolution of CFs without being trapped in local optima.

The niching method for CFs introduces the niching property to CF construction in XCS with the OF module, i.e. XOF. The niching property enables appropriate combinations of CFs to grow optimally complex CFs for hierarchical problems. This property accelerates the generalisation of XOF rules. With this new feature, XOF, as an extension of XCS, has the niching property in both the evolutions of rules and CFs.

Future research will consider extending XOF with the new features to a multi-agent system to target multitask learning. The availability of the OL as the representative patterns can facilitate the automation of transferring CFs among systems.



\bibliographystyle{splncs04}
\bibliography{genexity}

\end{document}